\pdfoutput=1

\documentclass[11pt]{article}

\usepackage[]{acl}
\usepackage{times}
\usepackage{latexsym}

\usepackage[T1]{fontenc}
\usepackage[utf8]{inputenc}

\usepackage{microtype}

\usepackage{color}
\usepackage{colortbl}
\usepackage{multirow}
\usepackage{booktabs}
\usepackage{graphicx}
\usepackage{amsmath}
\usepackage{float}

\title{Is the Computation of Abstract Sameness Relations\\ Human-Like in Neural Language Models?}

\author{Lukas Thoma \and Benjamin Roth \\
  Digital Philology \\
  Research Group Data Mining and Machine Learning \\
  University of Vienna, Austria \\
  \texttt{\{lukas.thoma, benjamin.roth\}@univie.ac.at} \\}

\begin{document}
\maketitle
\begin{abstract}
In recent years, deep neural language models have made strong progress in various NLP tasks. This work explores one facet of the question whether state-of-the-art NLP models exhibit elementary mechanisms known from human cognition. The exploration is focused on a relatively primitive mechanism for which there is a lot of evidence from various psycholinguistic experiments with infants. The computation of ``abstract sameness relations'' is assumed to play an important role in human language acquisition and processing, especially in learning more complex grammar rules. In order to investigate this mechanism in BERT and other pre-trained language models (PLMs), the experiment designs from studies with infants were taken as the starting point. On this basis, we designed experimental settings in which each element from the original studies was mapped to a component of language models. Even though the task in our experiments was relatively simple, the results suggest that the cognitive faculty of computing abstract sameness relations is stronger in infants than in all investigated PLMs. Our implementation is available on \url{https://github.com/lmthoma/computation_ASRs}
\end{abstract}

\section{Introduction}
It has been shown that deep learning language models (LMs) acquire general linguistic representations in pre-training which enable them to get fine-tuned for diverse NLP tasks \citep[e.g.,][]{rogers-etal-2020-primer}. Accordingly, language modeling approximates certain aspects of human language learning and processing. Thus, many research questions in deep learning NLP become more and more reminiscent of those known from grammar theory and psycholinguistics. This article builds on a specific set of insights from experimental linguistics and attempts to verify these findings for pre-trained language models (PLMs).

In psycholinguistics, artificial grammar learning approaches and related research fields on human cognition focus on the primitive core mechanisms behind the observable behavior in language tasks -- a certain number of neural circuits that in interaction generate performances that go beyond the sum of its parts \citep[e.g.,][]{endress20asb}. In BERTology, some efforts are already inspired by psycholinguistics \citep[e.g.,][]{ettinger-2020-bert}, however all of those we are aware of tackle rather high linguistic levels (e.g. syntax or semantics) which we consider a problematic starting point for several reasons. The most important one is that in NLP research, inevitably, one has to decide for a theory or formalism by which to study an LM -- and for higher levels of linguistic processing, theories are often quite complex and centers of heated debates. 

One important cognitive primitive for rule learning in humans is the detection of (abstract) sameness relations.\footnote{Sometimes also referred to as ``repetition rules'' or ``identity relations''.} It is assumed to be involved in complex grammar learning \citep[]{endress20asb}. A large number of experimental work on this phenomenon gathered a great deal of evidence that infants can easily learn abstract repetition rules \citep{rabagliati19tpo}, in terms of detecting sameness relations at an abstract level of representation, generalizing away from concrete examples. So far there are no studies in BERTology investigating the computation of abstract sameness relations (ASRs) in PLMs. Considering the high-level capabilities of the models, such as question answering, that are usually explored \citep[e.g.,][]{brown20gpt3}, this may not be surprising. However, in this paper we hypothesize that the computation of ASRs is one key primitive linguistic mechanism deep learning LMs acquire in pre-training. As this phenomenon is considered important for language learning and processing in humans, it may therefore explain one aspect of \textit{why} BERT and other transformer PLMs work so remarkably well.

In order to investigate whether PLMs compute ASRs in a human-like manner, we attempt to transfer the experimental designs from studies with infants as accurately as possible, so that each element of the source experiments maps to one component of state-of-the-art NLP models. In the underlying artificial grammar learning experiments \citep[e.g.,][]{marcus99rlb}, sequences of three syllables are used. Initially in a familiarization phase in which the sequences of a certain tri-gram syllable pattern are played to infants as an acoustic signal. For example, AAB, ABA, and ABB are tri-gram patterns with sameness relations, in which A and B are variables for syllables. An ABA structure sequence is played to infants, in order to prime them with this tri-gram pattern. To be more specific, if \textit{ga} and \textit{li} were A syllables, and \textit{ti} and \textit{na} were B syllables, the ABA familiarization tri-grams would be \textit{ga ti ga} and \textit{ga na ga}, as well as \textit{li ti li} and \textit{li na li}. In a later probing phase, tri-grams of syllables not used in the priming input are presented to infants, for example \textit{wo} and \textit{de} as A and \textit{fe} and \textit{ko} as B syllables. Based on observable behavior or neuroimaging methods, it is measured whether probe structures consistent with the priming phase (ABA in our example) are perceived differently from inconsistent ones (AAB and ABB). Since unfamiliar syllables are used in the probing phase, the detection of the primed sameness relations has to happen at an abstract representation level, which appears to be an easy tasks for infants -- already a few days after birth \citep{gervain08tnb}. 

To ensure that the investigated PLMs also process sequences consisting of exactly three elements (tri-grams), we use model tokens (representing subwords) instead of syllables in our experiments. Given a certain model input containing different instances of priming tri-grams with a particular kind of sameness relation (e.g. ABA), the probability a PLM assigns to a following probe tri-gram (either consistent or inconsistent with the sameness relation in the priming input) is calculated from the logits vector. Based on these probability values, a surprisal score $S(Probes|Primes)$ is calculated. This surprisal score would correspond to human surprise as measured by reaction time in psycholinguistic preferential looking paradigm experiments. Further details on the methodology are given in section \textit{\ref{ch:3} \nameref{ch:3}}.

This article attempts to demonstrate how experimental designs conducted with humans can be transferred into the domain of deep learning NLP research. Through mapping every relevant aspect of the source experiments to deep learning NLP models, we establish a direct relation and a clear expectation with respect to the results (if PLMs mirror human cognitive processing). The investigated state-of-the-art NLP models, however, could not fulfill this expectation. The results of our experiments indicate substantial differences in the computation of ASRs between humans and PLMs.

\section{Related Work}
This work contributes to research evaluating and improving deep learning NLP models based on what is known from human cognition. As cognition is a vast research field, and according to several linguistic theories, many aspects of it may be relevant for human language \citep[e.g.,][]{hauser02tfo, evans07tcl}, studies can be categorized in focusing on different levels. \citet{mcclelland20pli} start from a rather holistic perspective and attribute artificial neural networks utilizing query-based attention to rely on the same principles as the human mind: ``connection-based learning, distributed representation, and context-sensitive, mutual constraint satisfaction-based processing''. In their article they argue that future neural models of understanding should build equally on cognitive neuroscience and artificial intelligence, which is also the underlying idea in our research. There are many efforts in computational linguistics that address compositional generalization or the importance of structure in general which can also be categorized as rather higher level approaches to cognitive linguistics \citep{punyakanok08tio, poon09usp, collobert11nlp, lake18gws, li19cgf, russin19cgi, andreas20gcd, gordon19pem, akyurek20ltr, herzig20ssp, kim2020cogs, li20tnt, shaw20cga, conklin21mtc}. \citet{conklin21mtc} stand out by also considering the limits of human cognition -- based on insights from human intelligence research \citep{griffiths20uhi}. Thus, as in our paper, elementary concepts of human cognition -- the limits of working memory -- are used as a source of information to improve NLP performance, which ultimately leads to more robust generalizations in their work.

There is also a lot of relevant work around the computation of abstract sameness relations in humans, the elementary cognitive concept in our work. First and foremost there is \citet{marcus99rlb} and the behavioral experiments with infants which build the foundation of our experimental design. Furthermore, \citet{gervain08tnb} and \citet{kabdebon19sli} are to be mentioned as the most important follow-up studies influencing our efforts. In total, there are around 60 experiments on the computation of ASRs to date (with over 1,300 infants involved) which were all evaluated in a meta-analysis from \citet{rabagliati19tpo}. Drawing upon these efforts, there are several works that model the cognitive mechanisms underlying a sameness relation detection
\citep{arena13mti, luduena13asn, kumaran07wcm, grill06rat, wen08snd, hasselmo97fra, carpenter87amp, engel11sod, cope18acl, johnson09adn, endress20asb}.
\citet{endress20asb} is to be emphasized here. In his approach, biologically plausible mechanisms (disinhibitory neural net circuits) are introduced based on recent evidence from cognitive neuroscience and implemented as R computer models. The author points out that his approach is the first so far in which generalization to unseen stimuli does not require any kind of learning and therefore no negative evidence. Thus, the presented computer models show the behavior that is known from humans with respect to the computation of ASRs. A central assumption for us is that state-of-the-art deep learning LMs are technically capable of representing the required elementary mechanisms modeled in \citet{endress20asb}.\\

Another line of relevant studies investigate the syntactic faculties of LMs, all starting from rather higher-level grammar theories: In the subject area of understanding hierarchical structures in general \citep[]{kuncoro-etal-2018-lstms, linzen2018distinct, tang2018selfattention}, syntactic representations/embeddings \citep[]{lin2019open, liu2019linguistic, tenney2019bert, kim2020cogs}, syntax knowledge above the word level \citep[]{hewitt19asp, goldberg2019assessing, lin2019open}; as well as how models deal with specific syntactic phenomena, such as negative polarity items \citep[]{warstadt2019investigating}. The experiments presented in our article are not based on a natural language, therefore they fall into a line of work to generate artifical languages for studying deep neural language models \citep{bowman15tci, wang17tgd, ravfogel19sti, white21eti}. The most relevant works for us are those building on psycholinguistic methods. \citet{futrell19nlm} investigated the maintenance of syntactic state in several deep neural language models, drawing for example on \citet{levy11ple} who researched this phenomenon in humans. From a methodological point of view, the approach of \citet{ettinger-2020-bert}, who also draws upon human language experiments and aims to introduce a suite of psycholinguistic diagnostics for NLP models, is similar to the one of this paper: by analyzing output predictions in a controlled context (input), the language models do not need to be fine-tuned for a specific NLP task. Further evaluations that build upon psycholinguistic tests are 
\citet{linzen16ata}, 
\citet{chowdhury18rso},
\citet{gulordava18cgr},
\citet{marvin18tse}, and
\citet{wilcox18wdr} -- and all these analyses draw their conclusions based on the output probabilities of the language models, too. 

This paper complements the introduced efforts in starting from a very primitive level of language processing, at which there is less controversy in the fundamental (linguistic) theories than in the approaches mentioned here. The computation of ASRs is a phenomenon that can build on a strong evidence base, as well as on detailed modeling work that is biologically plausible, as shown by \citet{endress20asb}.

\section{Transferring experiments from infants to NLP models}\label{ch:3}

\begin{figure*}
    \centering
    \includegraphics[width=0.7\linewidth]{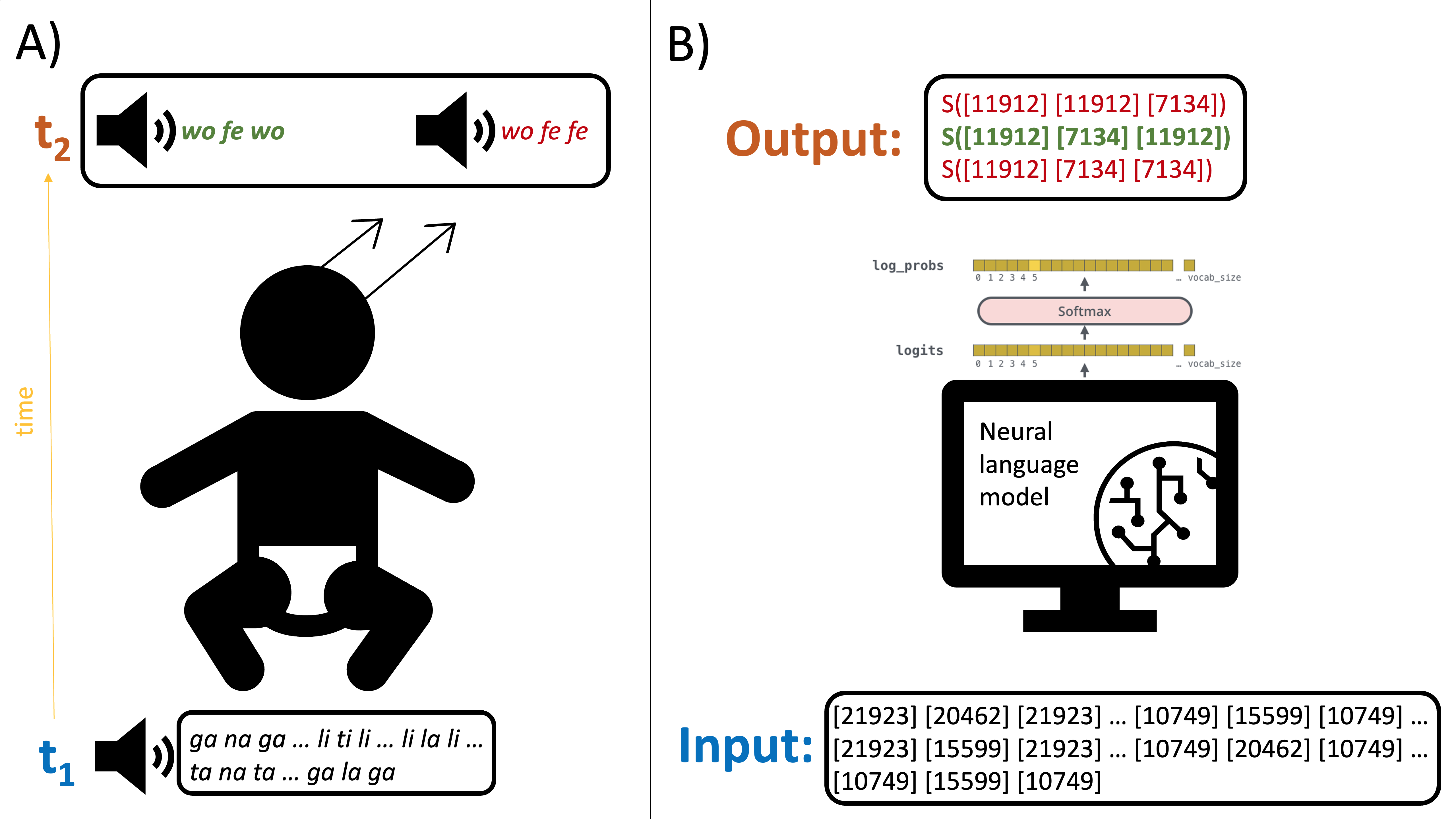}
    \caption{Experiments to investigate the computation of abstract sameness relations in humans and machines. A) shows a preferential looking paradigm setting with infants performed by, for example, \citet{marcus99rlb}: At $t_1$, syllable tri-grams with an ABA sameness relation are presented as audio signal. Later ($t_2$), the consistent ABA is played on the left speaker, the inconsistent ABB on the right speaker. As humans tend to focus on surprising percepts, looking longer into the direction of the speaker that plays the inconsistent stimuli indicates that the primed sameness relation (presented at $t_1$) was detected and abstracted to the consistent tri-gram that is build from unknown syllables. B) illustrates the basic experiment design in our work: Priming sequences (denoted as ``Primes'') act as model input upon which the surprisal scores elicited by a certain probe structure (``Probe'') are determined.}
    \label{fig:hVm}
\end{figure*}

The aim is to transfer one of the most prominent experimental designs for humans, \citet{marcus99rlb}, as accurately as possible, in order to determine whether PLMs are able to compute ASRs exactly in human-like manner. \textit{Figure \ref{fig:hVm}} shows a side-by-side illustration of the experiment, to better understand the respective correspondences. In general, the following key principles were pursued to achieve equivalence:

\begin{itemize}
    \item The prime and probe structures have to consist of exactly three perceived elements, as in the syllable sentences (tri-grams) for humans.
    \item A priming sequence not only contains several different instances of a structure (e.g. $n^2$ ABB sentences build from $n$ A and $n$ B elements) but also several tri-grams build from exactly the same elements (e.g. $m*n^2$ ABB sentences build from $n$ A and $n$ B elements). As in the experiments with infants, the tri-grams in the priming sequence are therefore not unique (e.g. $m$ times \textit{ga ti ti}, $m$ times \textit{li na na} and so on). The order of the tri-grams within a priming sequence is randomized.
    \item Specific elements (A and B) must not influence the structural evaluation by the subjects. In the original experiment, several strategies were followed to avoid statistical cues, such as controlling for phonetic features in the syllable material. 
\end{itemize}

In order to get the models operating internally with tri-grams, the priming and probing structures were build from tokens selected from the model vocabulary. Thus, tokens, as the smallest units of PLMs correspond to the syllables in the source experiments. At first, 2 A and 2 B prime tokens were randomly chosen, already assigned tokens were excluded from further selection. From this selection, 4 unique priming tri-grams per structure were generated, as shown in the following examples based on the subword sets $prime\ A=$ \{\textit{'river', 'shrill'}\} and $prime\ B=$ \{\textit{'hue', 'rt'}\}:

\begin{center}
\begin{tabular}{c c c}
    \# & \textbf{AAB} & \textbf{ABA}\\
    1 & \textit{'river river hue'} & \textit{'river hue river'}\\
    2 & \textit{'river river rt'} & \textit{'river rt river'}\\
    3 & \textit{'shrill shrill hue'} & \textit{'shrill hue shrill'}\\
    4 & \textit{'shrill shrill rt'} & \textit{'shrill rt shrill'}\\
\end{tabular}
\end{center}

\begin{center}
\begin{tabular}{c c c}
    \# & \textbf{ABB} & \textbf{ABC}\\
    1 & \textit{'river hue hue'} & \textit{'river hue shrill'}\\
    2 & \textit{'river rt rt'} & \textit{'river rt shrill'}\\
    3 & \textit{'shrill hue hue'} & \textit{'shrill hue river'}\\
    4 & \textit{'shrill rt rt'} & \textit{'shrill rt river'}\\
\end{tabular}
\end{center}

\noindent The data generation in the experiments of our work further follows those with infants, in that each unique tri-gram occurs several times in the priming sequence. Therefore the priming sequence always comprises 16 (4 times the 4 unique) tri-grams in randomized order. Analogously, 4 probe A and 4 probe B tokens are also selected from the model vocabulary and from these 16 unique tri-grams per structure are formed. Therefore, one experiment cycle comprised the model evaluation for 16 different probe tri-grams primed by the same sequence (4 times 4 unique tri-grams in randomized order), resulting in the following tri-gram sets:\\

\noindent\textbf{Primes:}
 \begin{itemize}
    \item $Prime\ Set\ AAB = \\ \{Prime\ AAB_1, Prime\ AAB_2, \\Prime\ AAB_3, Prime\ AAB_4\}$
    \item $Prime\ Set\ ABA = \\ \{Prime\ ABA_1, Prime\ ABA_2, \\Prime\ ABA_3, Prime\ ABA_4\}$
    \item $Prime\ Set\ ABB = \\ \{Prime\ ABB_1, Prime\ ABB_2, \\Prime\ ABB_3, Prime\ ABB_4\}$
\end{itemize}

\noindent \underline{Example priming sequence for $AAB\ primes$}:\footnote{The period (``.'') corresponds to a pause that was used between sentences in the audio priming input for infants.}\\
\textit{
shrill shrill hue.
river river rt.
river river hue.
river river hue.
river river rt.
river river rt.
river river rt.
shrill shrill rt.
shrill shrill rt.
shrill shrill hue.
river river hue.
shrill shrill hue.
river river hue.
shrill shrill hue.
shrill shrill rt.
shrill shrill rt.
}\\

\noindent\textbf{Probes:}
\begin{itemize}
    \item $Probe\ Set\ AAB = \\ \{Probe\ AAB_1, Probe\ AAB_2, ..., \\Probe\ AAB_{16}\}$
    \item $Probe\ Set\ ABA = \\ \{Probe\ ABA_1, Probe\ ABA_2, ..., \\Probe\ ABA_{16}\}$
    \item $Probe\ Set\ ABB = \\ \{Probe\ ABB_1, Probe\ ABB_2, ..., \\Probe\ ABB_{16}\}$
    \item $Probe\ Set\ ABC = \\ \{Probe\ ABC_1, Probe\ ABC_2, ..., \\Probe\ ABC_{16}\}$
\end{itemize}

In order to retrieve the probabilities from the model prediction, three different model inputs and thus inference times are required per priming-probing condition:\footnote{For simplicity, not all conditions are listed here; $Probe$ and $Primes$ denote all variants (AAB, ABA, ABB, ABC) that can be derived from the specified sets.} 
\begin{enumerate}
    \item $P_{t_1} = P(Probe_{p_1}|Primes)$
    \item $P_{t_2} = P(Probe_{p_2}|Primes,\ Probe_{p_1})$
    \item $P_{t_3} = P(Probe_{p_3}|Primes,\ Probe_{p_1},\ \\
    Probe_{p_2})$
\end{enumerate}

\noindent The subscript in $Probe$ refers to a position in the probe tri-gram, for example: $Probe_{p_1}$ = probe token at tri-gram position 1. 

As the selected token material must not distort the structural evaluation by the models, several measures were taken. Firstly, $P_{t_1}$ was not taken into consideration, because the probability of a random token given a certain priming sequence does not provide any information regarding the \textit{structural} prediction under investigation. Thus, we calculated a surprisal score based on the inverse log probabilities of $P_{t_2}$ and $P_{t_3}$:

\begin{equation}\label{f:422}
\begin{gathered}
 	S(Probe|Primes)\approx\\
    -\log_{2}P_{t_2}-\log_{2}P_{t_3}
\end{gathered}
\end{equation}

\noindent As shown in the example structures for AAB, ABA, ABB and ABC, we used the same subword sets $prime\ A$ and $prime\ B$ to generate all tri-grams. Thus, the features of randomly selected tokens impacted all structural evaluations in a similar way. In order to further reduce the impact of particular token selections, all surprisal scores presented in this paper are based on the mean value of 12288 different measurements per priming-probing condition: 16 different probe tri-grams per priming sequence (= experiment cycle), 256 different priming sequences per experiment run and 3 different experiment runs. As such large numbers are not feasible in experiments with infants, different strategies were pursued in the source studies. However, our results based on big randomized data can be considered equivalent, since both approaches aim at avoiding the impact of particular elements (tokens or syllables) distorting the overall structural evaluation by the subjects.

\begin{figure}[ht]
    \centering
    \includegraphics[width=.9\linewidth]{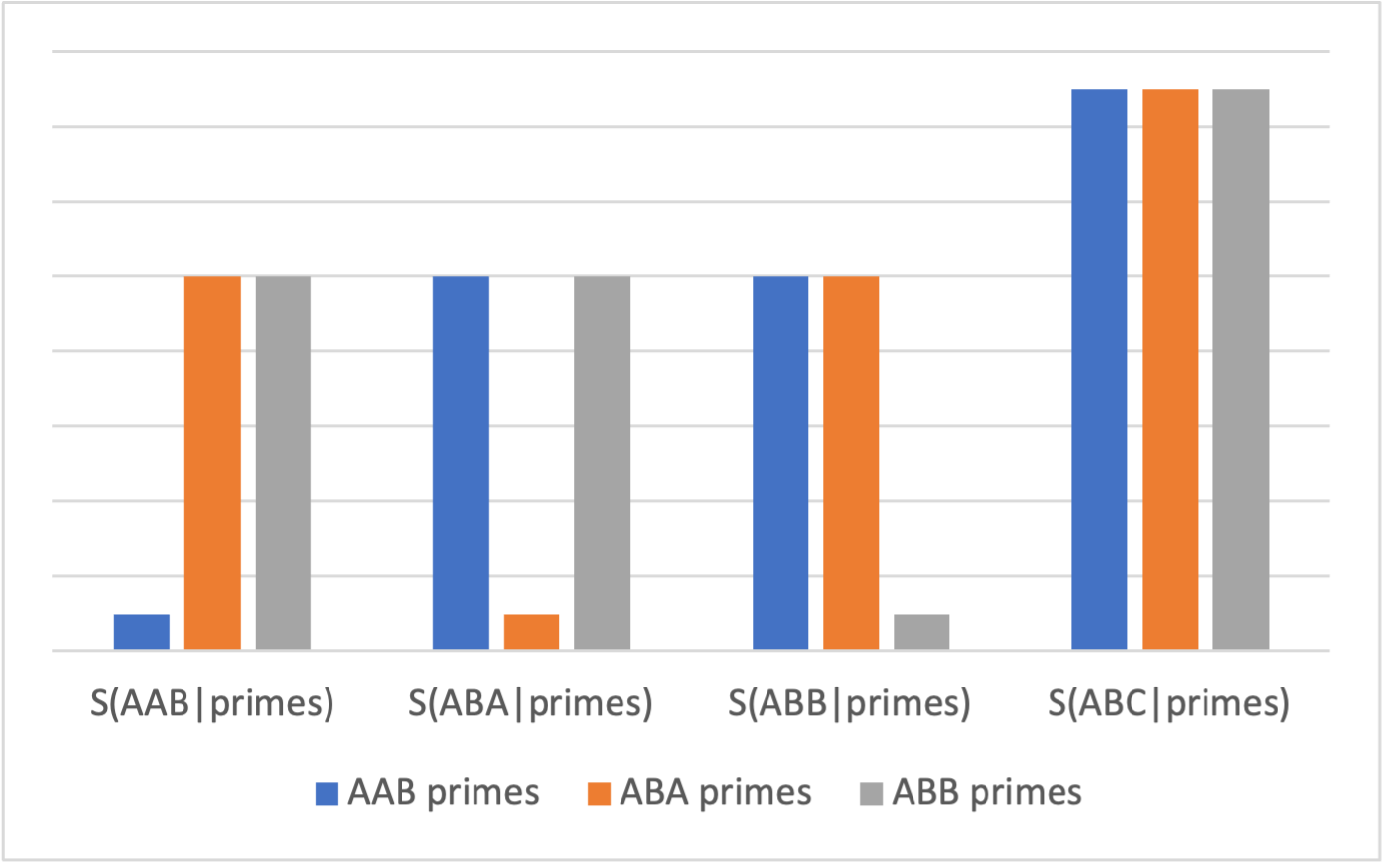}
    \caption{Schematic chart showing the expected distribution of surprisal scores for the investigated priming-probing conditions, based on the insights gathered from abstract sameness relation experiments with infants. The consistent conditions, e.g. for $ABB$ given $ABB\ primes$, evoke lower surprise as inconsistent ones. The surprisal scores for probing structures without sameness relations (e.g. $ABC|ABB\ primes$) are expected to be highest.}
    \label{fig:human}
\end{figure}

The reaction times measured in \citet{marcus99rlb} directly relate to the surprisal scores calculated in the presented way, therefore we \textit{expected} the values to be distributed as illustrated in \textit{Figure \ref{fig:human}} (since we hypothesized that the investigated LMs aquired a human-like computation of ASRs in pre-training).

A BERT model (bert-large-uncased), as well as a GPT-2 (gpt2) and an XLNet (xlnet-large-cased) with language modeling head from the (pre-trained) \textit{transformers} library \citep{wolf20huggingfaces} were investigated. The experiment design was the same for each PLM, with minor adaptations caused by the peculiarities regarding the model vocabulary and input format. Special tokens required in the model inputs (e.g. [CLS] or [MASK], as well as the token for the period) were excluded from the random prime and probe token selection.

\section{Results}
\subsection{Initial Results}

\begin{table*}[]
\begin{tabular}{llllll}
\textbf{} & \textbf{} & \textbf{S(AAB|primes)} & \textbf{S(ABA|primes)} & \textbf{S(ABB|primes)} & \textbf{S(ABC|primes)} \\ \hline
\textbf{BERT}  & AAB primes & \cellcolor[HTML]{D9E1F2}73.24 & 72.13                         & \textbf{70.27}                         & 74.31 \\
               & ABA primes & 71.38                         & \cellcolor[HTML]{FCE4D6}70.16 & \textbf{68.41}                         & 72.08 \\
               & ABB primes & 72.47                         & 71.09                         & \cellcolor[HTML]{D9D9D9}\textbf{69.41} & 73.18 \\ \hline
\textbf{XLNet} & AAB primes & \cellcolor[HTML]{D9E1F2}40.61 & 41.84                         & \textbf{40.17}                         & 43.48 \\
               & ABA primes & 40.77                         & \cellcolor[HTML]{FCE4D6}41.95 & \textbf{40.26}                         & 43.50 \\
               & ABB primes & 40.81                         & 41.88                         & \cellcolor[HTML]{D9D9D9}\textbf{40.26} & 43.45 \\ \hline
\textbf{GPT-2} & AAB primes & \cellcolor[HTML]{D9E1F2}57.65 & 57.62                         & \textbf{57.53}                         & 57.71 \\
               & ABA primes & 57.72                         & \cellcolor[HTML]{FCE4D6}57.69 & \textbf{57.60}                         & 57.77 \\
               & ABB primes & 57.66                         & 57.63                         & \cellcolor[HTML]{D9D9D9}\textbf{57.54} & 57.72 \\ \hline
\end{tabular}
\caption{Original Experiment: Random Primes, Random Probes}
\label{table:setting1}
\end{table*}

\begin{table*}[]
\begin{tabular}{llllll}
\textbf{} & \textbf{} & \textbf{S(AAB|primes)} & \textbf{S(ABA|primes)} & \textbf{S(ABB|primes)} & \textbf{S(ABC|primes)} \\ \hline
\textbf{BERT}  & AAB primes & \cellcolor[HTML]{D9E1F2}56.98 & 57.42                         & \textbf{55.89}                         & 59.00 \\
               & ABA primes & 56.56                         & \cellcolor[HTML]{FCE4D6}56.87 & \textbf{55.94}                         & 58.43 \\
               & ABB primes & 62.02                         & 62.20                         & \cellcolor[HTML]{D9D9D9}\textbf{60.86} & 63.92 \\ \hline
\textbf{XLNet} & AAB primes & \cellcolor[HTML]{D9E1F2}39.34 & 40.89                         & \textbf{39.08}                         & 42.55 \\
               & ABA primes & \textbf{41.28}                & \cellcolor[HTML]{FCE4D6}43.73 & 41.87                                  & 45.58 \\
               & ABB primes & 40.44                         & 41.81                         & \cellcolor[HTML]{D9D9D9}\textbf{40.22} & 42.70 \\ \hline
\textbf{GPT-2} & AAB primes & \cellcolor[HTML]{D9E1F2}57.09 & 57.07                         & \textbf{57.04}                         & 57.20 \\
               & ABA primes & 57.30                         & \cellcolor[HTML]{FCE4D6}57.28 & \textbf{57.23}                         & 57.34 \\
               & ABB primes & 56.13                         & 56.13                         & \cellcolor[HTML]{D9D9D9}\textbf{56.08} & 56.20 \\ \hline
\end{tabular}
\caption{Experimental Setting: Seen Primes, Random Probes}
\label{table:setting2}
\end{table*}

\begin{table*}[]
\begin{tabular}{lllll}
\textbf{} & \textbf{} & \textbf{S(AAB|primes)} & \textbf{S(ABA|primes)} & \textbf{S(ABB|primes)} \\ \hline
\textbf{BERT}  & AAB primes & \cellcolor[HTML]{D9E1F2}\textbf{44.67} & 50.87                         & 47.00                                  \\
               & ABA primes & \textbf{43.93}                         & \cellcolor[HTML]{FCE4D6}50.05 & 46.01                                  \\
               & ABB primes & \textbf{44.74}                         & 51.01                         & \cellcolor[HTML]{D9D9D9}47.02          \\ \hline
\textbf{XLNet} & AAB primes & \cellcolor[HTML]{D9E1F2}33.48          & 34.89                         & \textbf{33.46}                         \\
               & ABA primes & \textbf{33.33}                         & \cellcolor[HTML]{FCE4D6}34.80 & 33.44                                  \\
               & ABB primes & 33.52                                  & 34.83                         & \cellcolor[HTML]{D9D9D9}\textbf{33.44} \\ \hline
\textbf{GPT-2} & AAB primes & \cellcolor[HTML]{D9E1F2}\textbf{45.07} & 46.29                         & 46.82                                  \\
               & ABA primes & \textbf{45.12}                         & \cellcolor[HTML]{FCE4D6}46.34 & 46.87                                  \\
               & ABB primes & \textbf{45.06}                         & 46.30                         & \cellcolor[HTML]{D9D9D9}46.82          \\ \hline
\end{tabular}
\caption{Experimental Setting: Random Primes, Seen Probes}
\label{table:setting3}
\end{table*}

\begin{table*}[]
\begin{tabular}{lllll}
\textbf{} & \textbf{} & \textbf{S(AAB|primes)} & \textbf{S(ABA|primes)} & \textbf{S(ABB|primes)} \\ \hline
\textbf{BERT}  & AAB primes & \cellcolor[HTML]{D9E1F2}36.90          & 40.90                         & \textbf{36.83}                \\
               & ABA primes & \textbf{34.57}                         & \cellcolor[HTML]{FCE4D6}40.62 & 37.79                         \\
               & ABB primes & \textbf{46.79}                         & 48.93                         & \cellcolor[HTML]{D9D9D9}46.89 \\ \hline
\textbf{XLNet} & AAB primes & \cellcolor[HTML]{D9E1F2}\textbf{32.6}  & 36.89                         & 37.04                         \\
               & ABA primes & \textbf{32.13}                         & \cellcolor[HTML]{FCE4D6}33.73 & 33.76                         \\
               & ABB primes & 35.13                                  & \textbf{32.86}                & \cellcolor[HTML]{D9D9D9}34.78 \\ \hline
\textbf{GPT-2} & AAB primes & \cellcolor[HTML]{D9E1F2}\textbf{42.70} & 44.59                         & 44.87                         \\
               & ABA primes & \textbf{43.45}                         & \cellcolor[HTML]{FCE4D6}45.03 & 45.37                         \\
               & ABB primes & \textbf{42.05}                         & 43.70                         & \cellcolor[HTML]{D9D9D9}43.94 \\ \hline
\end{tabular}
\caption{Experimental Setting: Seen Primes, Seen Probes}
\label{table:setting4}
\end{table*}

Human-like behavior with respect to the computation of ASRs was clearly defined in \textit{Figure \ref{fig:human}}, however, as \textit{Table \ref{table:setting1}} shows, all models examined generated fairly unexpected results. The highlighted cells in this table mark the consistent priming-probing condition and thus the expected positions of the lowest surprisal scores. The actual minimum values are displayed in bold type, of which all can be found in the ABB probe column $S(ABB|primes)$. Therefore, ABB probing structures evoke the lowest surprisal score in all models, irregardless of the tri-gram structure presented in the priming sequence. Consequently, all investigated transformer models exhibited a very different behavior compared to the human subjects of the source experiments. The only expectation met was that the ABC probes elicited the highest mean values for the surprisal score in all priming conditions (cf. $S(ABC|primes)$ column). Considering the artificial language used in the experiments, the higher surprisal scores could originate in C being another nonsense token from the model perspective, whereas A and B are both already known from the input sequence at inference time 3. Therefore this part of the expected behavior is assumed to be only very indirectly related to a human-like computation of ASRs.\\

\subsection{\textit{pmi}-Based Tri-gram Selection}
Proceeding from our unexpected initial results, we performed additional experiments, based on facilitations in terms of ``seen data'' tri-grams. Since all models are pre-trained on extensive text corpora, they have already processed a number of token tri-grams with sameness relations. We supposed that a present ASR computation would rather be utilized, when the relevant tri-grams are not completely unknown to a PLM. In order to identify the already processed tri-grams with sameness relations, parts of the pre-training datasets were tokenized -- in accordance with the model under investigation. The BERT and XLNet models are, among others, pre-trained on the BooksCorpus dataset \citep{zhu15aba} which is freely available.\footnote{\url{https://huggingface.co/datasets/bookcorpus}} WebText, the dataset OpenAI GPT-2 is pre-trained on is not publicly available, however, \citet{gokaslan19openweb} created an ``open-source replication'' of it\footnote{\url{https://huggingface.co/datasets/openwebtext}} -- one third of WebText was scanned for relevant tri-grams. For all data sets, only tri-grams occurring at least 20 times were considered and for those, the Pointwise Mutual Information was calculated as follows:

\begin{equation}\label{f:431}
\begin{gathered}
 	pmi = \log_{2}(N^2* \\ \frac{C(3gram)}{(C(Tok_{p_1})*C(Tok_{p_2})*C(Tok_{p_3})})
\end{gathered}
\end{equation}

\noindent $N$ denotes the total number of tokens in the analyzed pre-training data subset, $C(3gram)$ the count of a respective tri-gram in the analyzed corpus, and $C(Tok_{p_1})$, $C(Tok_{p_2})$, and $C(Tok_{p_3})$ the count of the corresponding tri-gram token in the corpus at tri-gram position 1, 2, and 3, respectively. \textbf{From these \textit{pmi} results, we created a ranking of the top 32 tri-grams and from this ranking either the prime, the probe, or both tri-grams were selected.} When the prime tri-grams were selected from the $pmi$ ranking, all sentences in the priming sequence were unique, unlike as in the random condition. This was established deliberately, based on the idea that the model is more likely to recognize the sameness relation given several different instances of seen tri-grams with the same ASR. For random token priming sequences, the opposite is true, as more unknown (token or syllable) material contributes to more potential noise rather than helping to detect the underlying sameness relation pattern. The factor known vs. unknown data is assumed to invert this effect. Consequently, the seen data facilitations yielded three additional experimental settings : 1) Seen Primes, Random Probes, 2) Random Primes, Seen Probes, 3) Seen Primes, Seen Probes. 

As the results presented in \textit{Table \ref{table:setting2}}-\textit{Table \ref{table:setting4}} show, when random probe tri-grams are used in the experiments, ABB probes still cause the lowest surprisal score (with one exception for XLNet in the inconsistent AAB after ABA primes condition). The the majority of minima moves to the $S(AAB|primes)$ column when seen data tri-grams are used as probes. Thus, already processed tri-grams with sameness relations generate interesting effects, however, none of these point in the direction of a task facilitation -- PLMs still behave very differently compared to humans in computing ASRs in the conducted experiments. The intended facilitations could not be established by using seen data tri-grams with sameness relations.

\section{Discussion and Outlook}
Why all PLMs failed in this presumably easy task, cannot be sufficiently answered based on the experiments conducted so far. An ASR computation mechanism as known from experiments with infants (alone!) cannot explain the observed behavior of the PLMs. Further explorations are required in order to understand what happens at inference time -- whether a computation of ASRs is nonetheless available, however, for some reason not utilized in this kind of tasks. Analyses of attention mechanisms could reveal that a human-like computation of ASRs is indeed not present, or else, that it is interfered by other mechanisms with higher weights in the models. For the time being, it cannot be ruled out that the mechanism which enables infants to detect sameness relations at abstract levels is also present in PLMs.

It is further conceivable that the decision to work with tokens instead of text inputs, such as the same kind of syllable tri-grams used in the experiments with infants, might have been counterproductive, since tokens could play a subordinate role for PLMs in such tasks. Preliminary ad-hoc experiments with GPT-3 may support this hypothesis. On the OpenAI Playground\footnote{\url{https://beta.openai.com/playground}} (a web interface for GPT-3 and more) the following model input was given based on which the \textit{davinci} engine\footnote{Since this is the largest model available on this platform, we assume GPT-3 behind \textit{davinci}.} generated a correct output:
\begin{center}
\begin{tabular}{ll}
\textbf{Input} &
  \textbf{Output} \\
\textit{\begin{tabular}[c]{@{}l@{}}an ba : an ba an . \\ ka en : ka en ka .\\ pu ef : pu ef pu . \\ da ru :\end{tabular}} &
  \textit{\textbf{\begin{tabular}[c]{@{}l@{}}\\ \\ \\ \\ \\ \\ \ \ \ \ \ \ \ \ \ \ \ \ da ru da .\\ da re : da re da . \\ da se : da se da.\end{tabular}}}
\end{tabular}
\end{center}

With regard to these pilot experiments, it is also possible that capacity is an essential factor.\footnote{In this respect, the human brain should be superior for a few more decades, both in terms of the number of neurons and the connections between them \citep[e.g.,][Chapter 1.2.3]{goodfellow16dle}.} On the OpenAI Playground, when engines with fewer parameters were selected, an incorrect output was generated that did not contain tri-grams with sameness relations. The GPT-2 model used in the experiments, with 1.5 billion parameters, may not be able to acquire the computation of abstract sameness relations in pre-training, whereas the gigantic GPT-3 model, with 175 billion parameters, maybe is. Another hypothesis could be that computing ASRs might not be very relevant in processing English language, compared to, for example, Arabic where this mechanism is assumed to have essential functions in morpho-syntax \citep{endress09pam}. In an English-dominant pre-training, which applies for all investigated models, the computation of ASRs may therefore only be acquired, if sufficient capacity is available. For future research, it might be promising to include multilingual PLMs, since repetition rules are featured in the grammar of over 80 percent of languages. For English, this cognitive mechanism is assumed to be only a primitive building block for learning more complex grammar rules in the course of human language acquisition. So it is very likely not an explicit aspect in the data processed during model pre-training and since this primitive is not an ``innate'' circuit (as in humans), PLMs may develop alternative strategies to process English. 

As a final note, it should be considered that there are also multi-modal experimental designs with infants utilizing visual and auditory stimuli \citep{kabdebon19sli}. Coming back to the potentially inappropriate decision to use token tri-grams instead of plain text (e.g. similar syllables as used in the source experiments): Based on the multi-modal experiments with humans, it is reasonable to assume that the exact representation format or level of the variables (A or B) does not affect the basic function of the investigated primitive mechanism. As a consequence, the results of the so far conducted experiments suggest that PLMs are at least less flexible than humans with regard to the computation of abstract sameness relations, since they are not able to abstract away from token-level input. Apart from that, additional explorations are necessary to draw valid conclusions, which will be the starting point for our future research.

\section{Conclusion}
In this paper, methods from psycholinguistic research were transferred to the domain of deep learning NLP research to investigate an elementary cognitive mechanism that is considered central to rule-based grammar learning in cognitive linguistics: The computation of abstract sameness relations.

The results of the conducted experiments suggest that the computation of ASRs is clearly not human-like for the pre-trained BERT, XLNet and GPT-2 models examined. We did not expect these state-of-the-art language models having such problems with this allegedly simple task -- which infants can solve already a few days after birth. At this stage, however, our findings do not permit definitive conclusions. Nevertheless, the results provide insight and concrete suggestions for furhter experiments.

\bibliography{anthology,custom}
\bibliographystyle{acl_natbib}

\end{document}